\documentclass[letterpaper, 10 pt, conference]{ieeeconf}  

\IEEEoverridecommandlockouts                              

\usepackage{color}
\usepackage{amsmath}
\usepackage{amsfonts}
\usepackage{amssymb}
\usepackage{graphicx}
\usepackage{hyperref}
\usepackage{tabularx}

\title{\LARGE \bf EU Long-term Dataset with Multiple Sensors for Autonomous Driving}

\author{Zhi Yan$^{1}$, Li Sun$^{2}$, Tom{\'a}{\v s} Krajn{\'i}k$^{3}$, and Yassine Ruichek$^{1}$%
  \thanks{This work was supported by the Quality Research Bonus (BQR) of the University of Technology of Belfort-Montb\'eliard (UTBM), the Contrat de Plan \'Etat-R\'egion (CPER) 2015-2020 (Mobilitech), the CZ MSMT project (No. FR-8J18FR018) / PHC Barrande project (No. 40682ZH) (3L4AV), the OP VVV funded project CZ.02.1.01/0.0/0.0/16\_019/0000765 (Research Center for Informatics), the AdMaLL project in partnership with EPSRC FAIR-SPACE Hub (EP/R026092/1), and the NVIDIA GPU grant program.}%
  \thanks{$^{1}$CIAD UMR7533, Univ. Bourgogne Franche-Comté, UTBM, F-90010 Belfort, France
    {\tt\small \{zhi.yan, yassine.ruichek\}@utbm.fr}}%
  \thanks{$^{2}$Sheffield Robotics, University of Sheffield, UK
    {\tt\small li.sun@sheffield.ac.uk}}%
  \thanks{$^{3}$Artificial Intelligence Center, Czech Technical University, Czechia
    {\tt\small tomas.krajnik@agents.fel.cvut.cz}}%
}

\begin{document}

\maketitle
\thispagestyle{empty}
\pagestyle{empty}

\begin{abstract}
  The field of autonomous driving has grown tremendously over the past few years, along with the rapid progress in sensor technology.
  One of the major purposes of using sensors is to provide environment perception for vehicle understanding, learning and reasoning, and ultimately interacting with the environment.
  In this paper, we first introduce a multisensor platform allowing vehicle to perceive its surroundings and locate itself in a more efficient and accurate way.
  The platform integrates eleven heterogeneous sensors including various cameras and lidars, a radar, an IMU (Inertial Measurement Unit), and a GPS-RTK (Global Positioning System / Real-Time Kinematic), while exploits a ROS (Robot Operating System) based software to process the sensory data.
  Then, we present a new dataset (\url{https://epan-utbm.github.io/utbm_robocar_dataset/}) for autonomous driving captured many new research challenges (e.g. highly dynamic environment), and especially for long-term autonomy (e.g. creating and maintaining maps), collected with our instrumented vehicle, publicly available to the community.
\end{abstract}

\section{INTRODUCTION}
\label{sec:introduction}

Both academic research and industrial innovation into autonomous driving (AD) has seen tremendous growth in the past few years and is expected to continue to grow rapidly in the coming years.
This can be explained by two factors including, $1)$ the rapid development of hardware (e.g. sensors and computers) and software (e.g. algorithms and systems), and $2)$ the needs for travel safety, efficiency, and low-cost along with the development of human society.

\begin{figure}[t]
  \centering
  \includegraphics[width=\columnwidth]{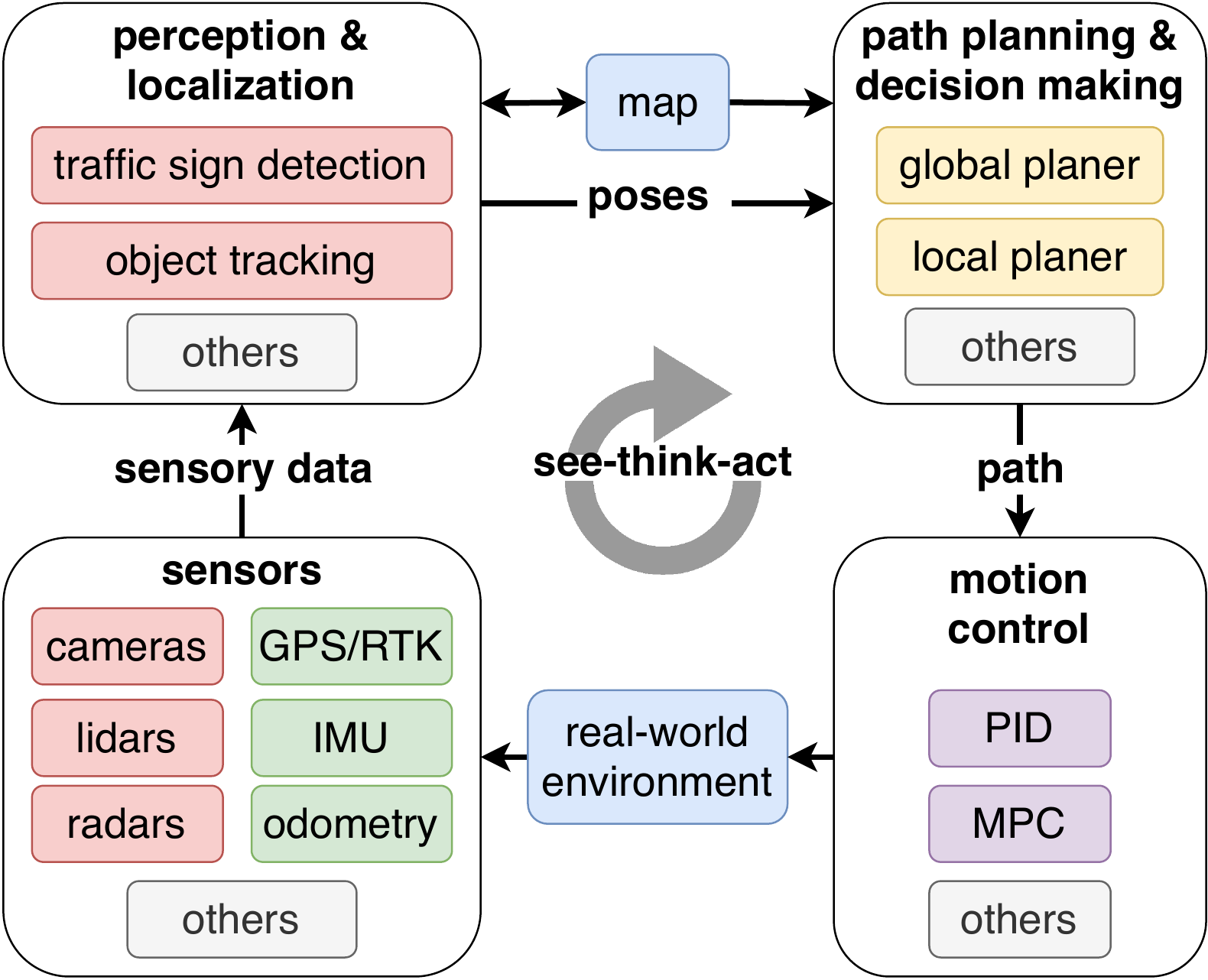}
  \caption{A general multisensor-based framework for a map-based AD system~\cite{siegwart}.}
  \label{fig:framework}
\end{figure}

A general framework for autonomous navigation of unmanned vehicle consists of four modules, including sensors, perception and localization, path planning and decision making, as well as motion control.
It's typically to have vehicles answer three questions: ``Where am I?'', ``What's around me?'', and ``What should I do?''.
As shown in Fig.~\ref{fig:framework}, the vehicle acquires the external environmental data (e.g. image, distance and velocity of object) and self-measurements (e.g. position, orientation, velocity and odometry) through various sensors.
Sensory data are then delivered to the perception and localization module, help the vehicle understand its surroundings and localize itself in a pre-built map.
Moreover, the vehicle is expected to not only understand what happened but also what is going on around it (e.g. for prediction)~\cite{fremen,ls18icra}, and it may simultaneously update the map with a description of the local environment for long-term autonomy~\cite{ls18ral,krajnik14icra}.
Afterwards, depending on the pose of the vehicle itself and other objects, a path is generated by the global planer and can be adjusted by the local planer according to the real-time circumstance.
Then the motion control module will calculate motor parameters to execute the path and send commands to the actuators.
Following the loop across these four modules, the vehicle can navigate autonomously via a typical see-think-act cycle.

Effective perception and localization are known as the most essential part of many modules for an autonomous vehicle to safely and reliably operating in our daily life.
The former includes the measurement of internal (e.g. velocity and orientation of the vehicle) and external (e.g. human, object and traffic sign) environmental information, while the latter mainly includes visual odometry / SLAM (Simultaneous Localization And Mapping), localization with a map, and place recognition / re-localization.
These two tasks are closely related and both affected by the sensors used and the processing manner of the data they provide.

Nowadays, the heterogeneous sensing system is commonly used in the field of robotics and autonomous vehicles in order to produce comprehensive environmental information.
Commonly used sensors include various cameras, 2D/3D lidar (LIght Detection And Ranging), radar (RAdio Detection And Ranging), IMU (Inertial Measurement Unit), and GNSS (Global Navigation Satellite System).
The combination use of these is mainly due to the fact that different sensors have different (physical) properties, and each category has its own pros and cons~\cite{yz18iros}.
On the other hand, ROS (Robot Operating System)~\cite{ros} has become the \emph{de facto} standard platform for development of software in robotics, and today increasing numbers of researchers and industries develop autonomous vehicles software based on it.
As an evidence, for example, seven emerging ROS-based AD systems were presented at ROSCon\footnote{\url{https://roscon.ros.org/}} 2017, while this number was zero in 2016.

In this paper, we report our progress in building an autonomous car at the University of Technology of Belfort-Montbéliard (UTBM) in France from September 2017, with a focus on the completed multisensor platform.
Firstly, we introduce a variety of sensors used for the purpose of efficient vehicle perception and localization, while illustrate the reason of choosing them, the installation positions, and some trade-offs we made in the system configuration.
Secondly, we present a new dataset for AD, entirely based on ROS, recorded with our platform in both urban and suburban areas, where all the sensors are calibrated, data are approximately synchronized (i.e. at the software level, except the two 3D lidars which are synchronized at the hardware level by communicating with positioning satellites), and the ground truth trajectories recorded by the GPS-RTK for vehicle localization is provided.
This dataset includes many new features for urban and suburban driving, such as highly dynamic environment (massive moving objects in vehicle odometry), roundabout, sloping road, construction bypass, aggressive driving, etc., and as it captures daily and seasonal changes, it is especially suitable for long-term vehicle autonomy research~\cite{kunze18ral}.
Moreover, we implemented the state-of-the-art methods as baselines for the lidar odometry benchmarking.
Finally, we illustrate the characteristics of the proposed dataset via a horizontal comparison with existing ones.

Getting started with autonomous vehicles might be a challenge and time consuming.
Because people have to face difficulties on the design, budgeting and cost control, and the implementation from the hardware (especially with various sensors) to the software level.
This paper is also expected to help readers quickly overcome similar problems through a comprehensive summary of our experience.
We hope these descriptions will give the community a practical reference.

\section{THE PLATFORM}
\label{sec:platform}

So far, there is no almighty and perfect sensor, and they all have limitations and edge cases.
For example, GNSS is extremely easy to navigate and works in all weather conditions, but its update frequency and accuracy are usually not enough to meet the requirements of AD.
Also, buildings and infrastructures in the urban environment are likely to obstruct the signals, thereby leading the positioning failures in many daily scenes such as urban canyons, tunnels, and underground parking lots.
Among visual and range sensors, the 3D lidar is generally very accurate and has a large field of view (FoV).
However, the sparse and geometry data (i.e. point clouds) obtained from this kind of sensors experience limited ability in semantic-related perception tasks.
In addition, in the case of vehicle traveling at high speed, relevant information is not handily extracted due to scan distortion (could be alleviated by motion compensation).
Furthermore, the lidar performance suffers from adverse weather conditions such as fog, rain, and snow~\cite{yang20iros,cadcd}.
The 2D lidar have obviously similar problems, with further limitations due the availability of a single scan channel and reduced FoV.
Nevertheless, 2D lidars are usually cheaper than the 3D ones, which have mature algorithm support and been widely used in mobile robotics long enough for mapping and localization problems.
Visual cameras can encode rich semantic and texture information into the image, while low robustness is experienced with lightness and illumination variances.
Radar is very robustness to light and weather changes, while it lacks of range sensing accuracy.

In summary, it is difficult to rely on a single sensor type for efficient perception and localization in AD, as concerned by this paper.
Hence, it is important for researchers and industries to leverage the advantages of different sensors and make the multisensor system complimentary with individual ones.
Table~\ref{tab:sensors_pros_and_cons} summarizes typical advantages and disadvantages of the commonly used sensors.

\begin{table}[t]
  \caption{Pros and cons of the commonly used sensors for AD}
  \label{tab:sensors_pros_and_cons}
  \begin{center}
    \begin{tabular}{|l|l|l|}
      \hline
      \textbf{Sensors} & \textbf{Pros} & \textbf{Cons}\\
      \hline\hline
      GNSS & easy-to-use & low positioning accuracy\\
      & less weather sensitivity & limited by urban area\\
      \hline
      lidar & high positioning accuracy & high equipment cost\\
      & fast data collection & high computational cost\\
      & can be used day and night & ineffective during rain\\
      \hline
      camera & low equipment cost & low positioning accuracy\\
      & providing intuitive images & affected by lighting\\
      \hline
      radar & reliable detection & low positioning accuracy\\
      & unaffected by the weather & slow data collection\\
      \hline
    \end{tabular}
  \end{center}
\end{table}

\subsection{Hardware}
\label{sec:hardware}

\begin{figure}[t]
  \centering
  \includegraphics[width=\columnwidth]{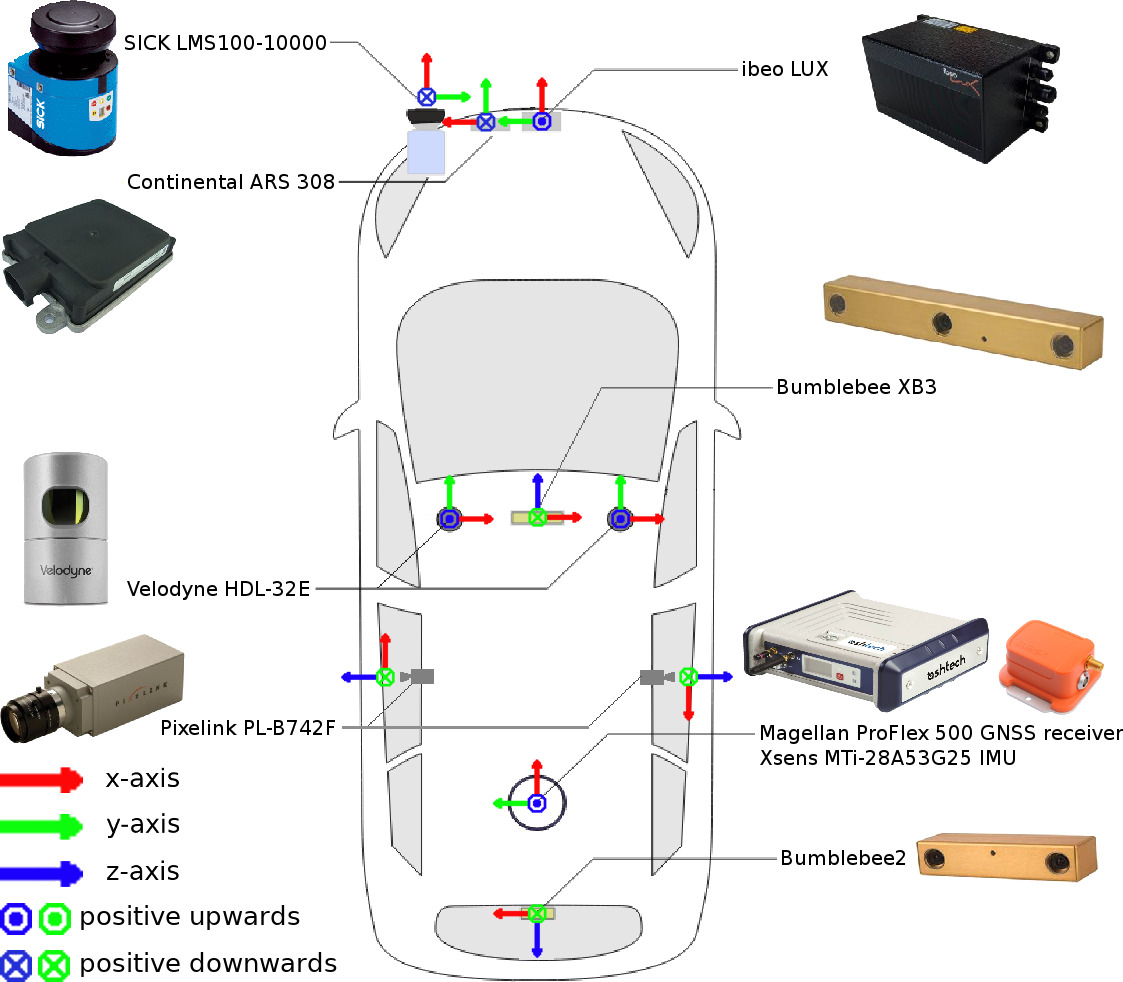}
  \caption{The sensors used and their mounting positions.}
  \label{fig:utbm_robocar_sensors}
\end{figure}

\begin{figure}[t]
  \centering
  \includegraphics[width=0.91\columnwidth]{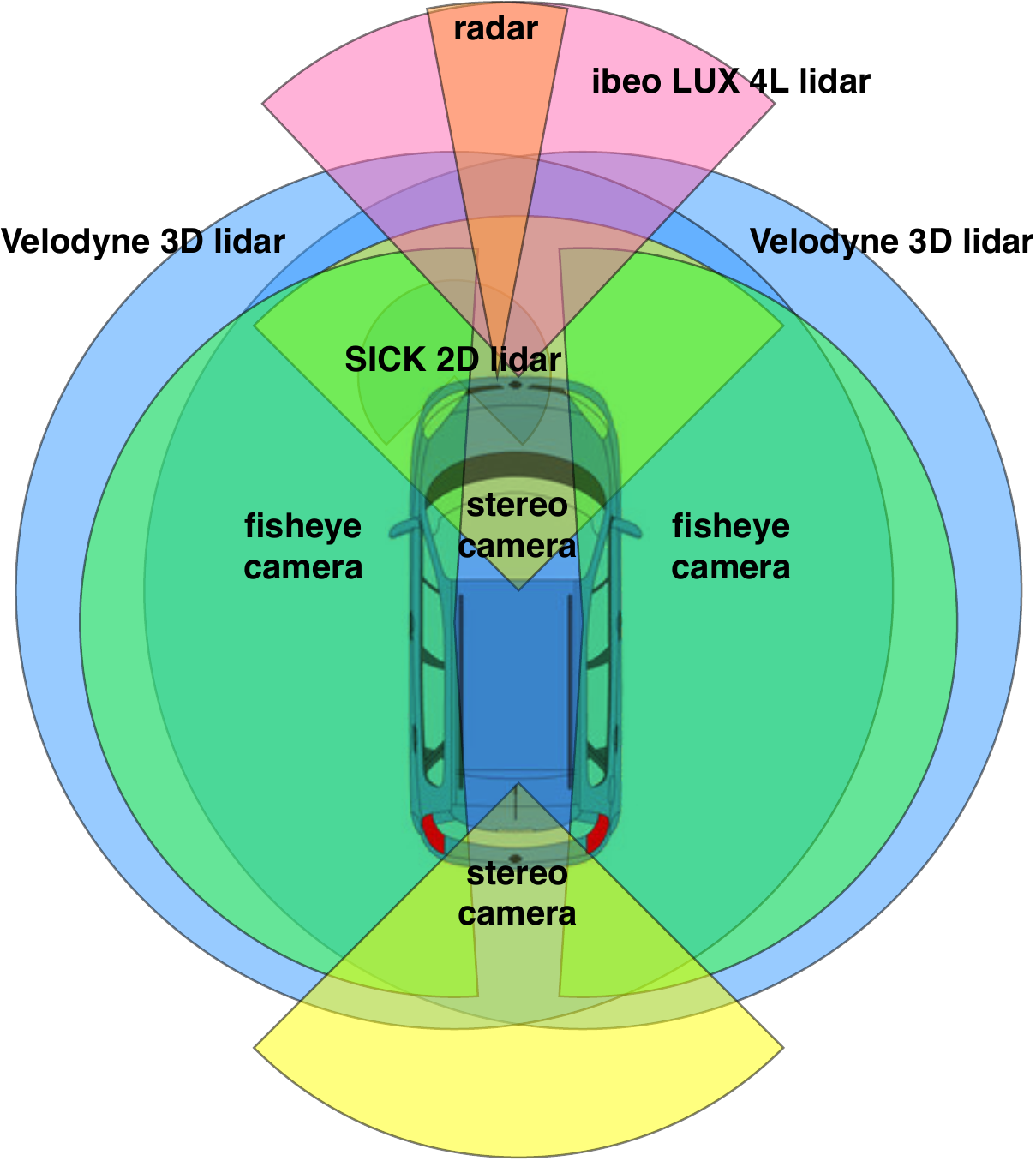}
  \caption{The visual scope of the vehicle sensors.}
  \label{fig:visual_scope}
\end{figure}

The sensor configuration of our autonomous car is illustrated in Fig.~\ref{fig:utbm_robocar_sensors}.
Its design for external environment perception mainly adheres to two principles (see Fig.~\ref{fig:visual_scope}): $1)$ strengthen the visual scope as much as possible, and $2)$ maximize the overlapping area perceived by multiple sensors.
In particular:
\begin{itemize}
\item Two stereo cameras, i.e. a front-facing Bumblebee XB3 and a back-facing Bumblebee2, are mounted on the front and rear of the roof, respectively.
  These two cameras are both with CCD (Charge-Coupled Device) sensors in global shutter mode, and compared to rolling shutter cameras, they are more advantageous when the vehicle is driving at a high speed.
  In particular, every pixel in a captured image is exposed simultaneously at the same instant in time in global shutter mode, while exposures typically move as a wave from one side of the image to the other in rolling shutter mode.
\item Two Velodyne HDL-32E lidars are mounted on the front portion of the roof, side by side.
  Each Velodyne lidar has 32 scan channels, 360$^{\circ}$ horizontal and 40$^{\circ}$ vertical FoV, with a reported measuring range up to 100m.
  It is noteworthy that when using multiple Velodyne lidars in proximity to one another, as in our case, sensory data may be affected due to one sensor picking up a reflection intended for another.
  In order to reduce the likelihood of the lidars interfering with each other, we used its built-in phase-locking feature to control where the laser firings overlap for the data recording, and post-processed it to remove data shadows behind each lidar sensor.
  Details will be given in Section~\ref{sec:configuration of two Velodyne lidars}.
\item Two Pixelink PL-B742F industrial cameras with Fujinon FE185C086HA-1 fisheye lens are installed in the middle of the roof, facing the lateral sides of the vehicle.
  The camera has CMOS (Complementary Metal-Oxide-Semiconductor) global shutter sensor that freezes the high-speed motion, while the fisheye lens allows to capture an extremely wide angle of view (185$^{\circ}$).
  This setting, on the one hand, increases the vehicle's perception of the environment on both lateral sides that has not been well studied so far, and on the other hand, adds a semantical complement to the Velodyne lidars.
\item An ibeo LUX 4L lidar is embedded into the front bumper close to the y-axis of the car, which provides four scanning layers, a 85$^{\circ}$ (or 110$^{\circ}$ if one uses only two layers) horizontal FoV, and up to 200m measurement range.
  Together with a radar, they are extremely important for our system to ensure the safety of the vehicle itself as well as other objects (especially humans) in the vicinity of the front of the vehicle.
\item A Continental ARS 308 radar is mounted in a position close to the ibeo LUX lidar, which is very reliable for the detection of moving objects (i.e. velocity).
  While less angularly accurate than lidar, radar can work in almost every condition and some models even use reflection to see behind obstacles~\cite{radar-through-wall}.
  Our platform is designed to detect and track objects in front of the car by ``cross-checking'' both radar and lidar data.
\item A SICK LMS100-10000 laser rangefinder (i.e. 2D lidar) facing the road is mounted on one side of the front bumper.
  It measures its surroundings in 2D polar coordinates and provides a 270$^{\circ}$ FoV.
  Due to its downward tilt, the sensor is able to scan the road surface and deliver information about road markings and road boundaries.
  The combination use of the ibeo LUX and the SICK lidars is also recommended by the industrial community, i.e. the former for object detection (dynamics) and the latter for road understanding (statics).
\item A Magellan ProFlex 500 GNSS receiver is placed in the car with two antennas on the roof.
  One antenna is mounted on the z-axis perpendicular to the car rear axle for receiving satellite signals and the other is placed at the rear of the roof for synchronizing with an RTK base station.
  With the help of the RTK enhancement, the GPS positioning will be corrected and the positioning error will be reduced from meters-level to centimeters-level.
\item An Xsens MTi-28A53G25 IMU is also placed inside the vehicle, putting out linear acceleration, angular velocity, absolute orientation, among others.
\end{itemize}

It is worth mentioning that a trade-off we made in our sensor configuration is the side-by-side use of two Velodyne 32-layer lidars rather than adopting a single lidar or other models.
The reason for this is twofold.
First, in the single lidar solution, the lidar is mounted on a ``tower'' in the middle of the roof in order to eliminate occlusions caused by the roof, which is not an attractive option from an industrial design point of view.
Second, other models such as 64-layer lidar is more expensive than two 32-layer lidars which cost more than two 16-layer lidars.
We therefore use a pair of 32-layer lidars as the trade-off between sensing efficiency and hardware cost.

Regarding the reception of sensory data, the ibeo LUX lidar and the radar are connected to a customized control unit that is used for real-time vehicle handling and low-level control such as steering, acceleration and braking.
This setting is very necessary, because the real-time response from these two sensors to CAN bus is extremely important for driving safety.
All the lidars via a high-speed Ethernet network, the radar via RS-232, the cameras via IEEE 1394, and the GPS/IMU via USB 2.0, are connected to a DELL Precision Tower 3620 workstation.
The latter is only for data collection purpose, while a dedicated embedded automation computer will be used as master computer ensuring operation of the most essential system modules such as SLAM, point cloud clustering, sensor fusion, localization, and path planning.
Then a gaming laptop (with high-performance GPU) will serve as slave unit which is responsible to process computational intense and algorithmically complex jobs, especially for the visual computing.
In addition, our current system is equipped with two 60Ah external car batteries that can provide us with more than one hour of autonomy.

\subsection{Software}
\label{sec:software}

\begin{figure}[t]
  \centering
  \includegraphics[width=\columnwidth]{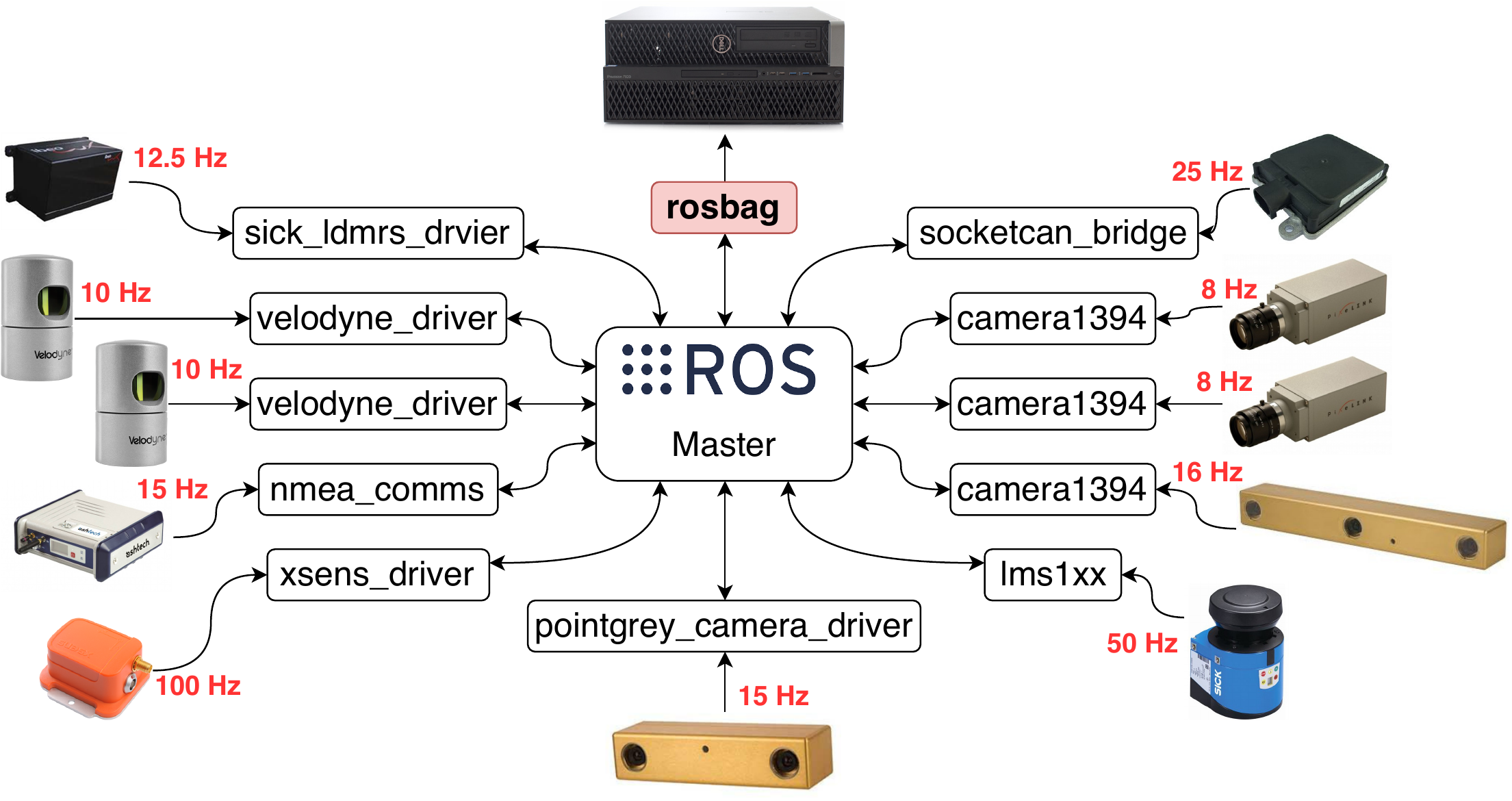}
  \caption{ROS-based software architecture diagram for data collection. The data is saved in \emph{rosbag} format. Please note that, in order to facilitate the reader to reproduce the system, we indicate the ROS package name instead of the ROS node name for each sensor driver. However, the ROS master communicates actually with the node provided by the package.}
  \label{fig:ros_graph}
\end{figure}

Our software system is based entirely on ROS.
For data collection, all the sensors are physically connected to the DELL workstation and all ROS nodes were running locally.
This setting maximizes data synchronization at the software level (timestamped by ROS)\footnote{Synchronization at the hardware level is beyond the scope of this paper.}.
The ROS-based software architecture diagram and the publish frequency of each sensor for data collection are shown in Fig.~\ref{fig:ros_graph}.
It is worth pointing out that the collection was done with a CPU-only (Intel i7-7700) computer, while no data delay was discovered.
This is mainly due to the fact that we only record the raw data and leave the post-processing to offline playback.
It is also worth noting that we focus on providing pioneering experience in vehicle perception purely based on ROS-1 (which can be a reference for ROS-2), and let loose the data collection at the vehicle regulation level.
Moreover, as we provide raw data from different devices, advanced processing such as motion compensation can be done by the end user.

\subsubsection{Sensor Calibration}

\begin{figure}[t]
  \centering
  \includegraphics[width=\columnwidth]{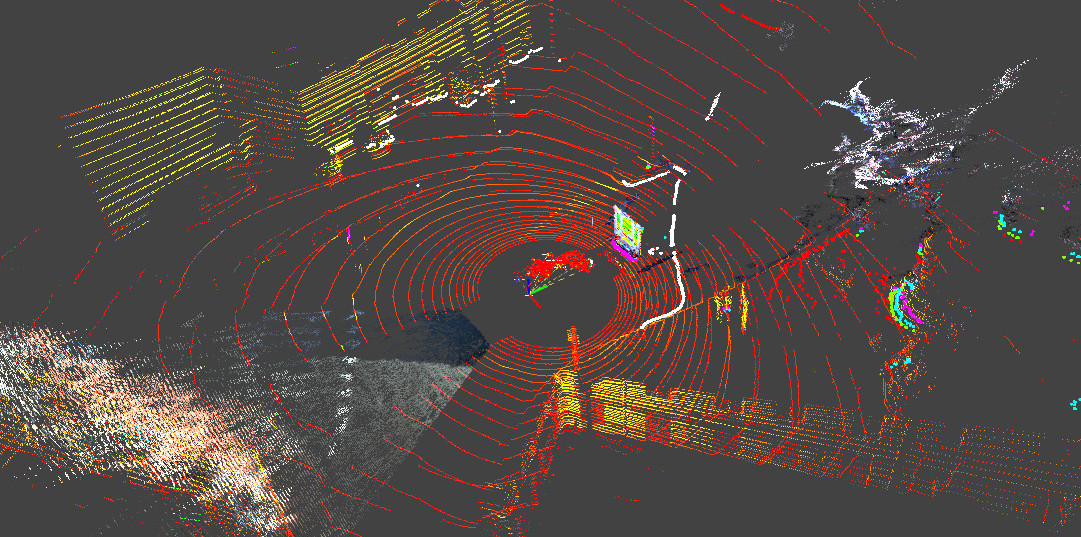}
  \caption{A ROS Rviz screenshot of the collected data with calibrated sensors. The autonomous car is in the centre of the image with a truck in front. The red ring points come from the Velodyne lidars, white points from the SICK lidar, and colored dots from the ibeo LUX lidar. The point clouds in front of and behind the car are from the two Bumblebee stereo cameras.}
  \label{fig:sensor_calibration}
\end{figure}

All our cameras and lidars were intrinsically calibrated, while the calibration files are available along with the dataset.
The calibration of the cameras were performed with a chessboard using ROS \emph{camera\_calibration} package, while the lidars are with factory intrinsic parameters.
The stereo cameras were also calibrated with respect to the Velodyne lidars.
The extrinsic parameters of the lidars were estimated via minimizing the voxel-wise $L2$ distance of the points from different sensors by driving the car in a structured environment with several landmarks.
To calibrate the transform between the stereo camera and the Velodyne lidar, we drove the car facing the corner of a building and manually aligned two point clouds on three planes i.e. two walls and the ground.
An aligned sensor data is visualized in Fig.~\ref{fig:sensor_calibration}.
It can be seen that through the calibration, points from all the lidars and the stereo cameras are aligned properly.

\subsubsection{Configuration of two Velodyne lidars}
\label{sec:configuration of two Velodyne lidars}

\begin{figure}[t]
  \centering
  \includegraphics[width=\columnwidth]{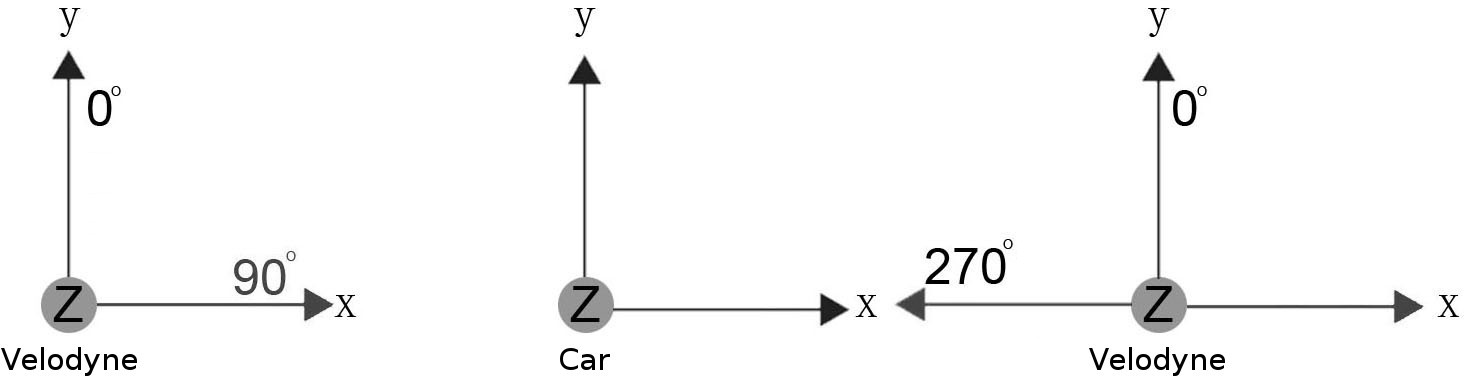}
  \caption{Phase offset setting of two side-by-side installed Velodyne lidars}
  \label{fig:phase_offset}
\end{figure}

As aforementioned, the two Velodyne lidars have to be properly configured in order to work efficiently.
Firstly, the phase lock feature of each sensor needs to be set to synchronize the relative rotational position of the two lidars, based on the Pulse Per Second (PPS) signal.
While the latter can be obtained from the GPS receiver connected to the lidar's interface box.
In our case, i.e. the two sensors are placed on the left and right sides of the roof, the left one has its phase lock offset set to 90$^{\circ}$, while the right one is set to 270$^{\circ}$, as shown in Fig.~\ref{fig:phase_offset}.

\begin{figure}[t]
  \centering
  \includegraphics[width=\columnwidth]{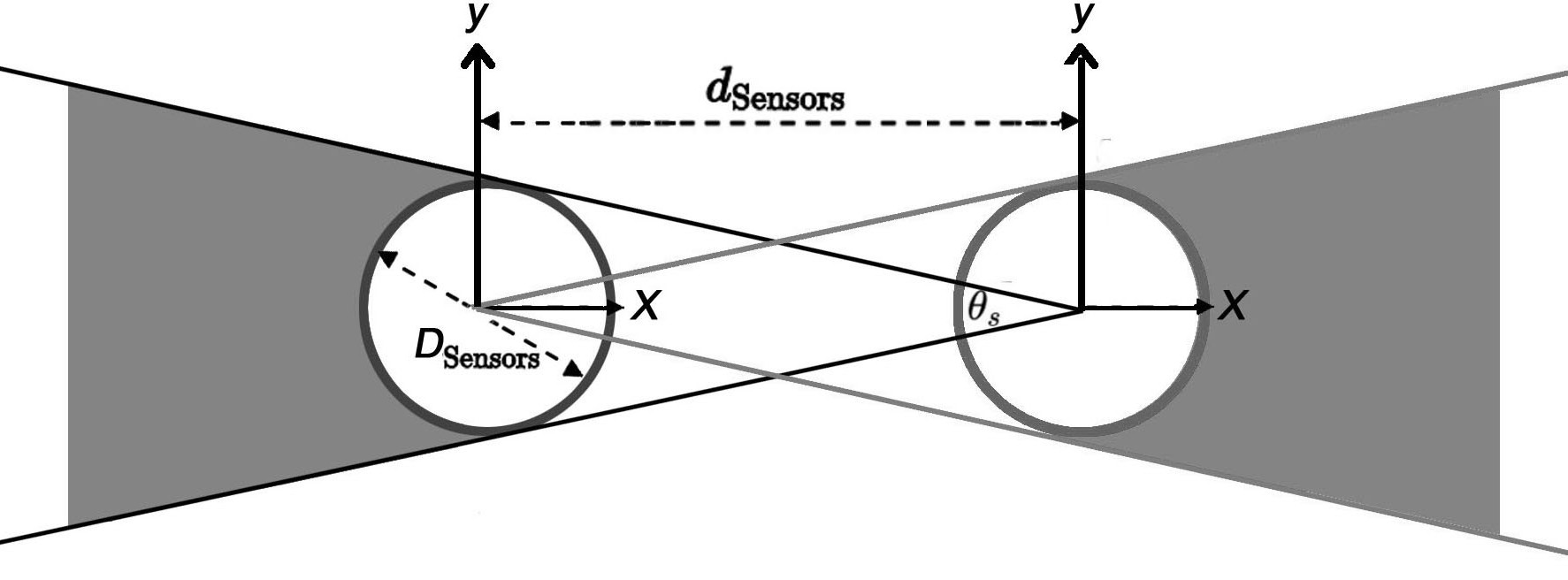}
  \caption{Data shadows behind a pair of Velodyne lidars.}
  \label{fig:data_shadows}
\end{figure}

Secondly, the Eq.~\ref{eq:data_shadows}~\cite{HDL-32E_User_Manual} can be used to remove any spurious data due to blockage or reflections from the opposing sensor (i.e. data shadows behind each other, see Fig.~\ref{fig:data_shadows}):
\begin{equation}
  \label{eq:data_shadows}
  \theta_s = 2 \times tan^{-1}(\frac{0.5 \times D_{sensor}}{d_{sensor}})
\end{equation}
where, $\theta_s$ is the subtended angle, $D_{sensor}$ is the diameter of the far sensor, and $d_{sensor}$ is the distance between sensor centers.

Moreover, in order to avoid network congestion led by the broadcast data of the sensors, we configure each Velodyne (the same for the SICK and the ibeo LUX lidars) to transmit its packets to a specific (i.e. non-broadcast) destination IP address (in our case, the IP address of the workstation), via a unique port.

\section{DATASET}
\label{sec:dataset}

\begin{figure}[t]
  \centering
  \includegraphics[width=0.47\columnwidth]{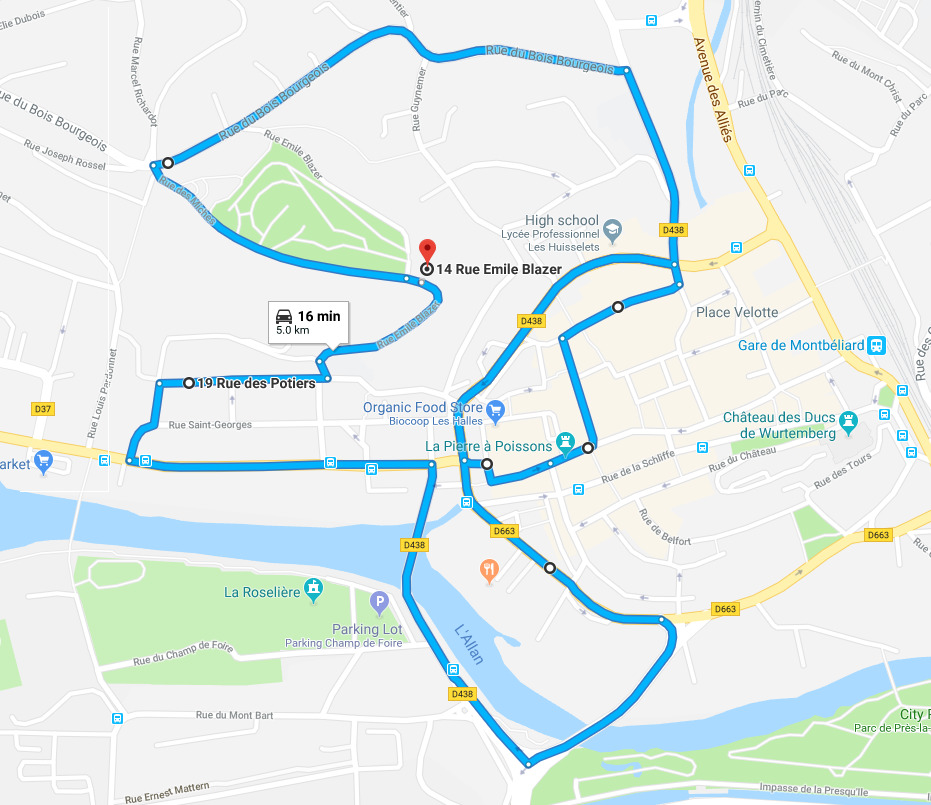}
  \includegraphics[width=0.502\columnwidth]{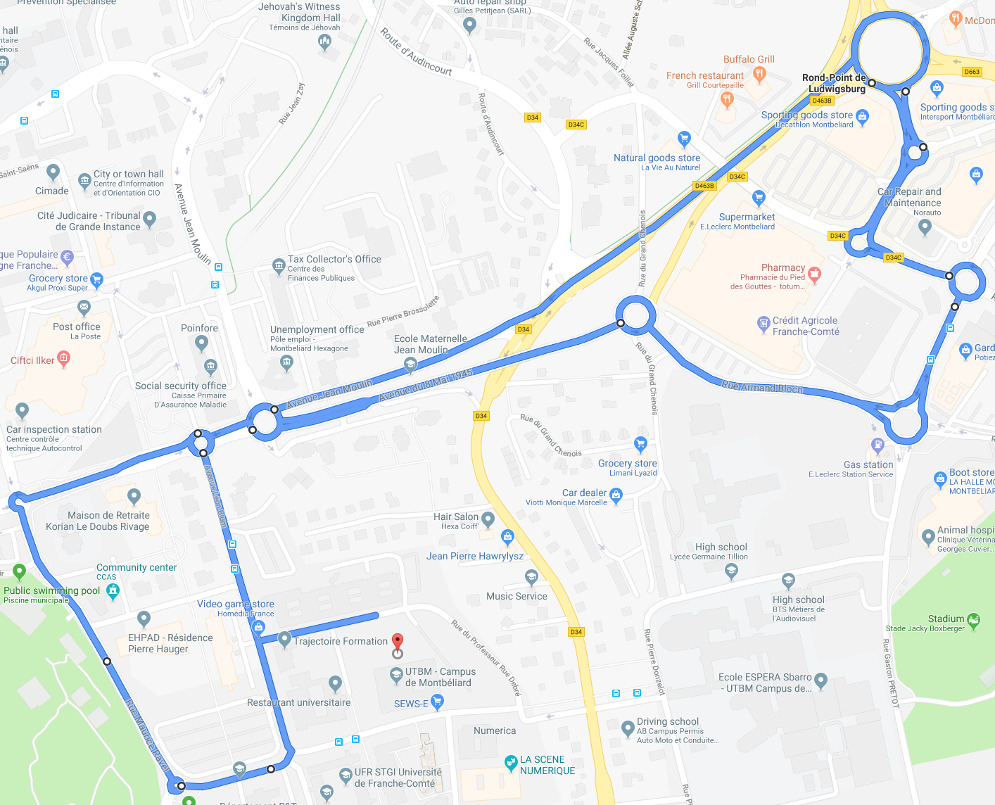}
  \caption{Data collection itineraries drawn on Google Maps. Left: for long-term data. Right: for roundabout data.}
  \label{fig:dataset_itinerary}
\end{figure}

Our recording software is fully implemented into the ROS system.
Data collection was carried out based on the Ubuntu 16.04 LTS (64-bit) and the ROS Kinetic.
The vehicle was driven by a human and any ADAS (Advanced Driver Assistant System) functions were disabled.
The data collection was performed in the downtown (for long-term data) and a suburb (for roundabout data) of Montb\'eliard in France.
The vehicle speed was limited to 50km/h following the French traffic rules.
It is conceivable that the urban scene during the day (recording time around 15h to 16h) was highly dynamic, while the evening (recording time around 21h) was relatively calm.
Light and vegetation (especially street trees) are abundant in summer, while winter is generally poorly lit, with little vegetation and sometimes even covered with ice and snow.
All data were recorded in \emph{rosbag} files for easy sharing with the community.
The data collection itineraries can be seen in Fig.~\ref{fig:dataset_itinerary}, which were carefully selected after many trials.

For the long-term data, we focus on the environment that is closely related to periodic changes~\cite{hypertime,vintr19icra} such as daily, weekly and seasonal changes.
We followed the same route eleven rounds at different times.
The length of the data recording is about 5km each round and the route passes through the city centre, a park, a residential area, a commercial area and a bridge on the river Doubs, and includes a small and a big road loop (for loop-closure purpose).
The RTK base station was placed at a fixed location on the mound - position marked by the red dot in Fig.~\ref{fig:dataset_itinerary}(left) (sea level 357m) - in order to communicate with the GNSS receiver in the car with minimal signal occlusion.
With these settings, we recorded data during the day, at night, during the week, in the summer and winter (with snow), always following the same itinerary.
At the same time, we captured many new research challenges such as uphill/downhill road, shared zone, road diversion, and highly dynamic/dense traffic.

Moreover, roundabouts are very common in France as well as in other European countries.
This road condition is not easy to handle even for humans.
The key is to accurately predict the behavior of other vehicles.
To promote related research on this topic, we repeatedly recorded some data in the area near the UTBM Montb{\'e}liard campus, which contains 10 roundabouts with various sizes in the range of approximately 0.75km$^2$ (see Fig.~\ref{fig:dataset_itinerary}(right)).

\subsection{Lidar Odometry Benchmarking}

As part of the dataset, we establish several baselines for lidar odometry\footnote{\url{https://github.com/epan-utbm/utbm_robocar_dataset}}, which is one of the challenges provided by our dataset.
We forked the implementation of the following state-of-the-art methods and experimented with our dataset:
\begin{itemize}
\item \emph{loam\_velodyne}~\cite{loam} is one of most advanced lidar odometry method and providing real-time SLAM for 3D lidar, submitted the state-of-the-art performance in KITTI benchmark~\cite{KITTI}.
  The implementation is robust for both structured (urban) and unstructured (highway) environments, and a scan restoration mechanism is devised for fast-speed driving.
\item \emph{LeGO-LOAM}~\cite{LeGO-LOAM} is a lightweight and ground-optimized LOAM, mainly to solve the problem that the performance of LOAM deteriorates when resources are limited and operating in noisy environments.
  Point cloud segmentation in LeGO-LOAM is performed to discard points that may represent unreliable features after ground separation.
\end{itemize}
Users are encouraged to evaluate their methods, compare with the provided baselines on devices with different levels of computation capability, and submit their results to our baseline GitHub repository.
However, only real-time performance is accepted, as it is critically important for the vehicle localization in AD.

\subsection{Long-term Autonomy}

Towards an on-the-shelf AD system, long-term autonomy, including long-term vehicle localization and mapping as well as dynamic object prediction, is necessary.
For this goal, we introduce the concept of ``self-aware localization'', ``liability-aware long-term mapping'' to advance the robustness of vehicle localization in a real-life and changing environment.
To be more specific, for the former, the vehicle should be empowered by global localisation technologies, e.g. global pose estimation~\cite{posenet} and loop closure detection~\cite{SegMatch}, to be able to wake up in any previously known locations.
While the ``liability-aware long-term mapping'' enables the vehicle to maintain the map in long-term with keeping the variance of landmarks updated and goodness of scan-map registration assessed~\cite{ls18ral}.
Moreover, the proposed long-term dataset can be used to predict occupancy and presence of dynamic objects such as humans and cars.
The periodical layout changes and human activities can be tracked and modelled using either frequency modelling~\cite{fremen} or Recurrent Neural Networks (RNNs)~\cite{ls18ral}.
The predicted occupancy map and human activity patterns can ultimately facilitate the vehicle motion planning in dynamic urban environments.
In this paper, we present the multiple sessions of driving data with a variance of lightness and landmarks, and propose the long-term localization and mapping as well as dynamic object prediction as open problems and encourage the researchers to investigate the potential solutions with our dedicated dataset.

\subsection{Roundabout Challenge}

Roundabout is unavoidable and can be very challenging for AD.
France has the largest number of roundabout in the world (about 50,000), with a considerable variety.
The various roundabout data we provide aims at pursuing related research on vehicle behavior prediction, and helping decreasing auto crashes in such situation.
On the one hand, one can get information about the car's turn signal from the image, and even the steering information of the wheels.
On the other hand, as we drove a full circle for each roundabout, users could have a long-term continuous data to learn and predict the trajectory of surrounding vehicles.

\section{RELATED WORK}
\label{sec:related_work}

Over the past few years, numerous platforms and resources for AD have emerged and grabbed public attention.
The AnnieWAY platform\footnote{\url{http://www.mrt.kit.edu/annieway/}} with its famous KITTI dataset\footnote{\url{http://www.cvlibs.net/datasets/kitti/}}~\cite{KITTI} have always shown strong influence in the community.
This dataset is the most widely-used visual perception dataset for AD, recorded with a sensing system comprising an OXTS RT 3003 GPS/IMU integrated system, a Velodyne HDL-64E 3D lidar, two Point Grey Flea 2 grayscale cameras, and two Point Grey Flea 2 color cameras.
With this configuration, the instrumented vehicle is able to produce 10 lidar frames per second with 100k points per frame for lidar based localization and 3D object detection, two gray images for visual odometry and two color images for optical flow estimation, object detection, tracking and semantic understanding benchmarks.

The RobotCar\footnote{\url{https://ori.ox.ac.uk/application/robotcar/}} from the University of Oxford is considered to be another powerful competitive platform.
The public available dataset\footnote{\url{https://robotcar-dataset.robots.ox.ac.uk/}}~\cite{Oxford} is the first multi-sensor long-term on-road driving dataset.
The Oxford RobotCar is equipped with a Bumblebee XB3 stereo camera, three Point Grey Grasshopper2 fisheye camera, two SICK LMS-151 2D lidar and a SICK LD-MRS 3D lidar.
Within this configuration, the three fisheye cameras cover a 360$^{\circ}$ FoV, the 2D/3D lidars and stereo cameras yield a data steam on 11fps and 16fps, respectively.
This dataset is collected in a period of one year and around 1000km in total.

KAIST dataset\footnote{\url{https://irap.kaist.ac.kr/dataset/}}~\cite{KAIST} focuses on complex urban environments such as downtown area, apartment complexes, and under-ground parking lot, and the data collection was performed with a vehicle equipped with 13 sensors.
Not long ago, Waymo\footnote{\url{https://waymo.com/open/}}~\cite{Waymo} (formerly the Google self-driving car project) started to release part of their data recorded across a range of conditions in multiple cities in the US.
More interesting is the recently released Canadian Adverse Driving Conditions Dataset\footnote{\url{http://cadcd.uwaterloo.ca/}}~\cite{cadcd}, which is designed to provide sensory data in varying degrees of snowfall.

Other datasets including
ApolloScape\footnote{\url{http://apolloscape.auto/scene.html}}~\cite{ApolloScape},
Cityscapes\footnote{\url{https://www.cityscapes-dataset.com/}}~\cite{Cityscapes} (collected with a stereo camera),
and BBD100K\footnote{\url{https://bair.berkeley.edu/blog/2018/05/30/bdd/}}~\cite{BBD100K} (collected with a monocular camera),
mainly focus on visual perception such as object detection, semantic segmentation, and lane/drivable area segmentation, and only visual data (i.e. images and videos) are released.
As the present paper focuses more on multisensor perception and localization, we do not give further details of these datasets here.
To have a more intuitionistic view, a comparison between our dataset and the existing ones is provided in Table~\ref{tab:comparison}.

\begin{table*}[t]
  \caption{A comparison of the existing datasets for AD}
  \label{tab:comparison}
  \begin{center}
    \begin{tabularx}{\textwidth}{|X|l|l|l|l|l|l|}
      \hline
      \textbf{Dataset} & \textbf{Sensor} & \textbf{Synchronization} & \textbf{Ground-truth} & \textbf{Location} & \textbf{Weather} & \textbf{Time}\\
      \hline\hline
      Ours & 2$\times$ 32-layer lidar & software & GPS-RTK/IMU & France$^1$ & sun, clouds, & day, dusk, night,\\
      & 1$\times$ 4-layer lidar & (ROS timestamp) & for vehicle & & snow & three seasons \\
      & 1$\times$ 1-layer lidar & and hardware & self-localization & & & (spring, summer,\\
      & 2$\times$ stereo camera & (PPS for the & & & & winter)\\
      & 2$\times$ fisheye camera & two Velodynes) & & & & \\
      & 1$\times$ radar & & & & &\\
      & 1$\times$ GPS-RTK & & & & &\\
      & 1$\times$ independent IMU & & & & &\\
      \hline
      KITTI~\cite{KITTI} & 1$\times$ 64-layer lidar & software & scene flow, odometry & Germany$^1$ & clear & day, autumn\\
      & 2$\times$ grayscale camera & and hardware & object detection & & & \\
      & 2$\times$ color camera & (reed contact) & \& tracking, & & & \\
      & 1$\times$ GPS-RTK/IMU & & road \& lane & & & \\
      \hline
      Oxford~\cite{Oxford} & 1$\times$ 4-layer lidar & software & GPS-RTK/INS & UK$^2$ & sun, clouds, & day, dusk, night,\\
      & 2$\times$ 1-layer lidar & & for vehicle & & overcast, rain & four seasons\\
      & 1$\times$ stereo camera & & self-localization & & snow &\\
      & 3$\times$ fisheye camera & & & & &\\
      & 1$\times$ GPS-RTK/INS & & & & &\\
      \hline
      KAIST~\cite{KAIST} & 2$\times$ 16-layer lidar & software & SLAM algorithm & South Korea$^1$ & clear & day\\
      & 2$\times$ 1-layer lidar & (ROS timestamp) & for vehicle & & &\\
      & 2$\times$ monocular camera & and hardware & self-localization & & &\\
      & 1$\times$ consumer-level GPS & (PPS for the & & & &\\
      & 1$\times$ GPS-RTK & two Velodynes, & & & &\\
      & 1$\times$ fiber optics gyro & an external trigger & & & &\\
      & 1$\times$ independent IMU & for the two & & & &\\
      & 2$\times$ wheel encoder & monocular cameras & & & &\\
      & 1$\times$ altimeter & to get stereo) & & & &\\
      \hline
      ApolloScape~\cite{ApolloScape} & 2$\times$ 1-layer lidar$^3$ & unknown & scene parsing, & China$^1$ & unknown & day\\
      & 6$\times$ monocular camera & & car instance, & & &\\
      & 1$\times$ GPS-RTK/IMU & & lane segmentation, & & &\\
      & & & self localization, & & &\\
      & & & detection \& tracking, & & &\\
      & & & trajectory, stereo & & &\\
      \hline
      nuScenes~\cite{nuScenes} & 1$\times$ 32-layer lidar & software & HD map-based & US$^1$ & sun, clouds, & day, night\\
      & 6$\times$ monocular camera & & localization, & Singapore$^2$ & rain & \\
      & 5$\times$ radar & & object detection & & & \\
      & 1$\times$ GPS-RTK & & \& tracking & & &\\
      & 1$\times$ independent IMU & & & & &\\
      \hline
      Waymo~\cite{Waymo} & 5$\times$ lidar$^4$ & unknown but very & object detection & US$^1$ & sun, rain & day, night\\
      & 5$\times$ camera$^4$ & well-synchronized & \& tracking & & &\\
       \hline
      CADC~\cite{cadcd} & 1$\times$ 32-layer lidar & hardware & object detection & Canada$^1$ & snowfall & day\\
      & 8$\times$ monocular camera & & \& tracking & & &\\
      & 1$\times$ GPS-RTK/IMU & & & & &\\
      & 2$\times$ independent IMU & & & & &\\
      & 1$\times$ ADAS kit & & & & &\\
      \hline
    \end{tabularx}
  \end{center}
  \begin{center}
    \begin{tabularx}{\textwidth}{|X|l|l|l|l|l|l|l|}
      \hline
      \textbf{Dataset} & \textbf{Distance} & \textbf{Data format} & \textbf{Baseline}$^5$ & \textbf{Download} & \textbf{License} & \textbf{Privacy} & \textbf{First release}\\
      \hline\hline
      Ours & 63.4km & rosbag (All-in-One) & 2 & free & CC BY-NC-SA 4.0 & face \& plate & Nov. 2018 \\
      & & & & & & removed &\\
      \hline
      KITTI~\cite{KITTI} & 39.2km & bin (lidar), png (camera) & 3 & registration & CC BY-NC-SA 3.0 & removal & Mar. 2012\\
      & & txt (GPS-RTK/IMU) & & & & under request &\\
      \hline
      Oxford~\cite{Oxford} & 1010.46km & bin (lidar), png (camera) & 0 & registration & CC BY-NC-SA 4.0 & removal & Oct. 2016\\
      & & csv (GPS-RTK/INS) & & & & under request &\\
      \hline
      KAIST~\cite{KAIST} & 190,989km & bin (lidar), png (camera) & 1 & registration & CC BY-NC-SA 4.0 & removal & Sep. 2017\\
      & & csv (GPS-RTK/IMU) & & & & under request &\\
      \hline
      ApolloScape~\cite{ApolloScape} & unknown & png (lidar), jpg (camera) & 1 & registration & ApolloScape License & removal & Apr. 2018\\
      & & & & & & under request &\\
      \hline
      nuScenes~\cite{nuScenes} & 242km & xml & 3 & registration & CC BY-NC-SA 4.0 & face \& plate & Mar. 2019\\
      & & & & & & removed &\\
      \hline
      Waymo~\cite{Waymo} & unknown & range image (lidar) & 3 & registration & Waymo License & face \& plate & Aug. 2019\\
      & & jpeg (camera) & & & & removed &\\
      \hline
      CADC~\cite{cadcd} & 20km & bin (lidar), png (camera) & 0 & registration & CC BY-NC 4.0 & removal & Jan. 2020\\
      & & txt (GPS-RTK/IMU/ADAS) & & & & under request &\\
      \hline
    \end{tabularx}
  \end{center}
  $^1$right-hand traffic,
  $^2$left-hand traffic,
  $^3$vertical scanning,
  $^4$device model undisclosed,
  $^5$only including methods published with the paper, excluding community contributions.
\end{table*}

For a deeper analysis, KITTI provides a relative comprehensive challenges for both perception and localization, and its hardware configuration, i.e. a combination of 3D lidar and stereo cameras, is widely-used for prototyping robot cars by autonomous vehicle companies.
While, there are still two limitation of KITTI dataset.
First, the dataset only captured in one session and long-term variances, e.g. lightness, season, of the scene are not investigated.
Second, the visual cameras have not covered the full FoV, thereby blind spots existed.
Oxford dataset investigated the vision based perception and localization with variance of seasons, weather and time, however, the modern 3D lidar sensory data is not included.
In this paper, we leverage the pros of the platform design in KITTI and Oxford, and eliminate the cons.
That is, a combination of four lidars (including two Velodynes) and four cameras multisensor platform is proposed to engender stronger range and visual sensing.

Other emerging datasets have also demonstrated strong competitiveness.
Waymo provides well synchronized and calibrated high quality LiDAR and camera data that are also exhaustively annotated.
This production-grade dataset will undoubtedly make a significant contribution to the community.
KAIST provides an alternative solution to multiple lidar platforms, i.e. two 16-layer lidars are mounted on both sides of the roof at an angle of 45$^{\circ}$ to maximize data acquisition coverage, and two 1-layer lidars are mounted on the rear and front of the roof facing downwards and upwards.

Apart from the hardware configuration and dataset collection, there exist widely-cited open-source repositories, such as
Apollo\footnote{\url{https://github.com/ApolloAuto/apollo}},
Autoware\footnote{\url{https://github.com/CPFL/Autoware}},
and Udacity\footnote{\url{https://github.com/udacity/self-driving-car}},
which provide researchers a platform to contribute and share AD software.

\section{CONCLUSION}
\label{sec:conclusion}

In this paper, we presented a dataset for AD research and the multisensor platform used for data collection.
The platform integrates eleven heterogeneous sensors including various lidars and cameras, a radar, and a GPS/IMU, in order to enhance the vehicle's visual scope and perception capability.
By exploiting the heterogeneity of different sensory data (e.g. sensor fusion), the vehicle is also expected to have a better localization and situation awareness, and ultimately improve the safety of AD for human society.

Leveraging our instrumented car, a ROS-based dataset is cumulatively recorded and is publicly available to the community.
This dataset is full of new research challenges and as it contains periodic changes, it is especially suitable for long-term autonomy research such as persistent mapping~\cite{ls18ral}, long-term prediction~\cite{vintr19icra,ls18icra}, and online/lifelong learning~\cite{fremen,yz18iros,hypertime,yz19auro,yz17iros}.
We hope our efforts and on-the-shelf experience could pursue the development and help on solving related problems in AD.

Furthermore, as we take privacy very seriously and handle personal data in line with the EU's data protection law (i.e. the General Data Protection Regulation (GDPR)), we used deep learning-based methods\footnote{\url{https://github.com/epan-utbm/image_anonymization}} to post-process the camera-recorded images in order to blur face and license plate information.
The images have been released successively from the first quarter of 2020.


\bibliographystyle{IEEEtran}
\bibliography{references}

\end{document}